\documentclass[pdflatex,sn-mathphys-ay,iicol]{sn-jnl}

\usepackage{graphicx}
\usepackage{multirow}
\usepackage{amsmath,amssymb,amsfonts}
\usepackage{amsthm}
\usepackage{mathrsfs}
\usepackage[title]{appendix}
\usepackage{xcolor}
\usepackage{textcomp}
\usepackage{manyfoot}
\usepackage{booktabs}
\usepackage{algorithm}
\usepackage{algorithmicx}
\usepackage{algpseudocode}
\usepackage{listings}
\usepackage{array}
\usepackage{url}
\usepackage{stfloats}

\begin{document}

\title[EEG-Based Prediction of Music Enjoyment]{EEG-Based Prediction of Music Enjoyment Using Deep Learning with Explainable Temporal Analysis}

\author[1]{\fnm{Parshav} \sur{Singla}}\email{psingla1\_be22@thapar.edu}
\author*[1]{\fnm{Simranjit} \sur{Kaur}}\email{simranjit.kaur@thapar.edu}
\author[1]{\fnm{Maanya} \sur{Walia}}\email{mwalia\_be22@thapar.edu}
\author[1]{\sur{Niharika}}\email{nniharika\_be22@thapar.edu}

\affil*[1]{\orgdiv{Department of Computer Science and Engineering},
\orgname{Thapar Institute of Engineering and Technology},
\city{Patiala}, Punjab, 147004, \country{India}}

\abstract{Understanding why listeners enjoy music is difficult because the experience is subjective and changes as a piece unfolds. Electroencephalography (EEG) offers a non-invasive way to study neural responses during listening, yet earlier work has mostly treated music emotion as a categorical label or relied on black-box models. We formulate enjoyment prediction as a regression problem on the public MUSIN-G dataset, which contains 128-channel EEG from 20 listeners together with song-level enjoyment ratings. Six deep learning models were compared under a participant-wise protocol, in which 15 listeners were used for training and validation and 5 were held out for testing. DeepConvNet, CNN-LSTM and GNN-CNN reached the lowest mean absolute error (0.97 to 0.99 on the 1 to 5 scale), while the temporal convolutional network (TCN) had the highest Pearson correlation with the ratings (r = 0.21). Gradient-based and occlusion-based explanations, averaged over all held-out windows, identify the time segments within each two-second window that most influence the TCN predictions. The study provides a reproducible cross-listener benchmark for EEG-based enjoyment regression and shows how explainability analysis complements standard metrics.}

\keywords{Electroencephalography, Music enjoyment, Deep learning,
Temporal modeling, Explainable AI, Regression}

\maketitle

\section{Introduction}\label{sec1}

Music enjoyment is a deeply personal and temporally evolving experience,
heavily influenced by rhythm, melody, familiarity, and individual
preference. Understanding how these acoustic features translate into
cortical activity remains an open challenge for affective computing and
cognitive neuroscience. Electroencephalography (EEG) provides a
valuable, non-invasive window into these neural dynamics as listeners
process music. However, decoding subjective enjoyment directly from
EEG data is notoriously difficult. Factors such as inherent signal noise,
substantial differences in brain activity across individuals, and the
lack of explicit ground-truth labels at a fine temporal scale make this a
highly complex task~\citep{Miyapuram2022}.

Recently, advances in deep learning have dramatically improved our ability
to model EEG data by learning spatial and temporal patterns end-to-end.
Convolutional architectures excel at mapping spatial dependencies across
the scalp, while temporal models are highly effective at capturing
sequential neural dynamics as they unfold over
time~\citep{Lawhern2018EEGNet,Schirrmeister2017DeepConv}. Despite these
advances, the vast majority of music-related EEG research remains limited
to predicting broad categorical emotions or classifying rhythms. These
studies often rely on randomized trial-level data splits that artificially
inflate how well the model generalizes to new
listeners~\citep{Wang2023Survey,Ma2024Review}. Furthermore, complex deep
learning approaches are frequently treated as black boxes, offering little
insight into exactly how specific neural patterns relate to the listener's
actual experience of music.

In this work, we address these limitations by formulating EEG-based music
enjoyment as a continuous regression problem, rigorously evaluated under a
stringent participant-wise protocol. Using the MUSIN-G dataset, which
contains high-density EEG recordings from 20 participants with
corresponding self-reported enjoyment ratings~\citep{Miyapuram2022}, we
systematically benchmark convolutional, temporal, and hybrid neural
architectures. Crucially, to move beyond simple predictive metrics, we
integrate explainable artificial intelligence (XAI) techniques to analyze
the temporal relevance of EEG segments. Our goal is not just to
build an accurate predictive model, but to uncover how sustained neural
dynamics correlate with the subjective human experience of music enjoyment.

The main contributions of this paper are as follows: (1) Music enjoyment
is formulated as a continuous regression task under a strict
participant-wise protocol emphasizing cross-subject generalization. (2)
Six architectures spanning convolutional, temporal, recurrent, and
graph-based designs are systematically compared under identical conditions.
(3) Explainability is treated as a core analytical goal through
gradient-based, saliency-based, and occlusion-based XAI methods. 

\section{Related Work}\label{sec2}

\subsection{EEG-Based Affective Computing}

The study of EEG-based music perception has a rich history in affective
computing. Early neurophysiological research focused heavily on spectral
analysis of canonical EEG bands during music listening, discovering clear
links between neural oscillations and emotional engagement mapped to
arousal and valence spaces~\citep{Russell1980Circumplex,Koelstra2012DEAP}.
Traditional machine learning pipelines later utilized handcrafted features
for music emotion recognition, but these methods frequently struggled to
generalize across different subjects~\citep{Alarcao2019Survey}.

Deep learning approaches have since transformed the field \citep{Zheng2015SEED}. Convolutional
Neural Networks (CNNs) have enabled robust hierarchical feature learning
directly from multichannel EEG
signals~\citep{Schirrmeister2017DeepConv,Lawhern2018EEGNet}, while hybrid
architectures were developed to explicitly capture temporal dependencies.
More recently, Temporal Convolutional Networks (TCNs) have shown
exceptional performance in capturing long-range temporal structures using
dilated convolutions with stable
optimization~\citep{Ingolfsson2020TCNet}. Graph Neural Networks (GNNs)
have also been explored for modeling complex functional connectivity
between EEG channels, although their robustness in cross-subject
evaluations remains
challenging~\citep{Song2020GNN}.

\subsection{Music Enjoyment and Preference Modeling}

While categorical emotion recognition has been extensively researched,
the neural signatures underlying subjective music enjoyment and preference
are distinct and often obscured by standard affective models. Foundational
studies by \citet{Lin2010EEGMusic} and by \citet{Hadjidimitriou2012Liking} were among the first to
show the feasibility of classifying music liking through time-frequency
analysis. Furthermore, recent evidence suggests that inter-subject neural
synchrony strongly correlates with sustained musical engagement ~\citep{Madsen2019Synchrony}. Despite these
advances, highly subjective constructs like enjoyment are still frequently
discretized into simple categories in the EEG
literature, stripping away the nuanced temporal variation of the listening
experience~\citep{Wang2023Survey,Ma2024Review}. This study directly bridges
this gap by prioritizing continuous, regression-based enjoyment prediction
supported by robust explainability.

\section{Dataset and Problem Formulation}\label{sec3}

The datasets used in this study are publicly available and were accessed
in accordance with their respective usage and licensing policies.

\textbf{MUSIN-G Dataset.}
The Music Listening--Genre EEG dataset (MUSIN-G) was collected at the
Indian Institute of Technology Gandhinagar and is publicly available at
OpenNeuro~\citep{Miyapuram2022}. It provides continuous EEG recordings
from 20 participants listening to music across diverse genres, accompanied
by self-reported familiarity and enjoyment ratings.

\subsection{MUSIN-G Dataset Details}

The MUSIN-G dataset contains recordings from 20 participants who listened
to 12 song excerpts of about two minutes each spanning multiple genres. EEG data were
recorded using a 128-channel system under eyes-closed conditions to
minimize ocular artifacts. Following each song, participants rated their
overall enjoyment on a standard 1--5 Likert scale~\citep{Miyapuram2022}.

\subsection{Regression Formulation}

Let $X \in \mathbb{R}^{C \times T}$ denote an EEG segment featuring $C$
channels and $T$ temporal samples, and let $y \in [1,5]$ denote the
participant's corresponding enjoyment rating. The objective is to learn a
mapping $f_\theta(X) = \hat{y}$, where $\hat{y}$ accurately approximates the ground-truth enjoyment score.
Ratings are normalized to $[0,1]$ during the training phase and remapped
to the original 1--5 scale during evaluation. To prevent subject-level
data leakage, a strict participant-wise split (15 participants for
training and validation, 5 completely held out for testing) is enforced
across all evaluations~\citep{Brookshire2024Leakage}.

\section{Methodology}\label{sec4}

\subsection{EEG Preprocessing}

\begin{figure}[ht]
\centering
\includegraphics[width=\columnwidth]{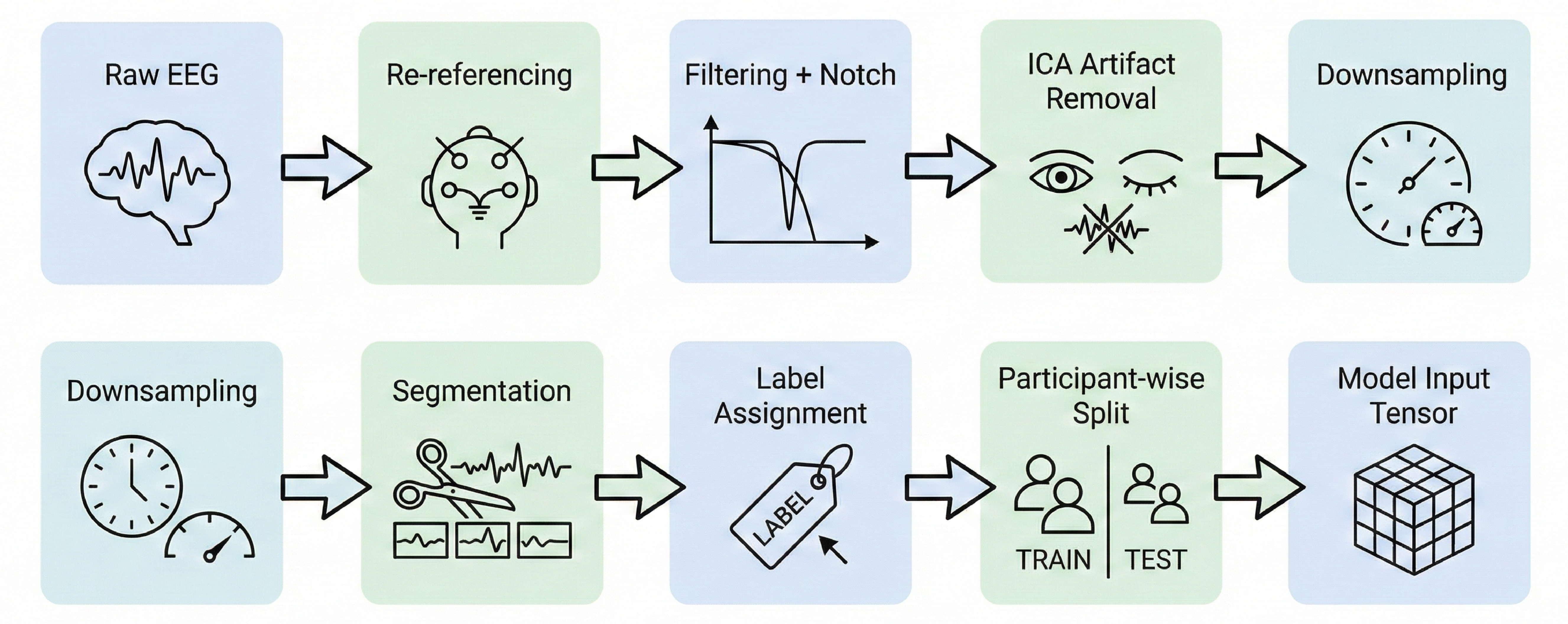}
\caption{EEG preprocessing and analysis pipeline, showing average referencing, high-pass filtering, ICA-based artifact rejection, downsampling, channel interpolation, segmentation, and participant-wise data splitting}
\label{fig:pipeline}
\end{figure}

Raw EEG signals are first average-referenced and high-pass filtered at
0.2\,Hz. The data are then cleaned using Independent Component Analysis
(ICA), applying MARA-based artifact rejection via
EEGLAB. The cleaned data are downsampled to
250\,Hz, bad channels are interpolated via spherical splines, and
continuous recordings are segmented into fixed-length, non-overlapping
windows of 500 samples. Because enjoyment labels are provided per song,
all windows derived from the same song inherit the same label. This
weak-label approach is effective for temporal modeling, but introduces
some label noise as human enjoyment naturally fluctuates across a full
song~\citep{Wang2023Survey}. Figure~\ref{fig:pipeline} illustrates the
complete preprocessing pipeline.

\subsection{Loss Function and Optimization}

Models are optimized using a combined regression loss that balances
sensitivity to large prediction errors with robustness against minor
outliers, as formulated in Equation \ref{eq:loss}:
\begin{equation}
\mathcal{L} = \alpha \cdot \mathcal{L}_{\text{MSE}}
            + (1 - \alpha) \cdot \mathcal{L}_{\text{MAE}},
\label{eq:loss}
\end{equation}
where $\alpha = 0.6$. The individual components are standard Mean Squared Error and Mean Absolute Error evaluations between ground truth ($y$) and prediction ($\hat{y}$).

The models are optimized using Adam and AdamW optimizers with an initial
learning rate of $3 \times 10^{-4}$, utilizing CosineAnnealing scheduling
for stable convergence. Dropout rates of 0.5--0.6 and early stopping
after 5--7 epochs of no validation improvement are applied. The full
training configuration is listed in Table~\ref{tab:training_params}.

\begin{table}[ht]
\centering
\caption{Training hyperparameters applied consistently across all models}\label{tab:training_params}
\begin{tabular}{@{}ll@{}}
\toprule
\textbf{Parameter} & \textbf{Value / Description} \\
\midrule
Batch size                & 16 \\
Initial learning rate     & $3\times10^{-4}$ \\
Optimizer                 & Adam / AdamW \\
LR scheduler              & CosineAnnealing \\
Dropout rate              & 0.5--0.6 \\
Maximum epochs            & 50 \\
Early stopping patience   & 5--7 epochs \\
Loss function             & $0.6\,\mathcal{L}_{\text{MSE}} + 0.4\,\mathcal{L}_{\text{MAE}}$ \\
Input tensor shape        & $128 \times 500$ \\
Target normalization      & Normalized to $[0,1]$ \\
Evaluation split          & 15 train / 5 test \\
\bottomrule
\end{tabular}
\end{table}

\subsection{Evaluation Metrics}

Four complementary metrics are used for performance assessment. Mean
Absolute Error (MAE) measures average absolute error. Root Mean Squared
Error (RMSE) penalizes larger errors more strongly. The Pearson correlation coefficient
measures trend alignment. Tolerance accuracy counts the percentage of predictions within $\pm 1$ of the true rating. Table~\ref{tab:metrics} summarizes these metrics.

\begin{table}[ht]
\centering
\caption{Evaluation metrics and their purpose in enjoyment regression}\label{tab:metrics}
\begin{tabular}{@{}lll@{}}
\toprule
\textbf{Metric} & \textbf{Direction} & \textbf{Description} \\
\midrule
MAE            & Lower is better & Avg. absolute deviation \\
RMSE           & Lower is better & Penalizes large errors \\
Pearson $r$    & Higher is better & Trend alignment \\
Tol. Accuracy & Higher is better & Predictions within $\pm 1$ \\
\bottomrule
\end{tabular}
\end{table}

\section{Models Evaluated}\label{sec5}

To ensure a comprehensive analysis, a diverse set of deep learning
architectures was benchmarked under the same preprocessing pipeline and
evaluation protocol.

\textbf{EEGNet}~\citep{Lawhern2018EEGNet} serves as a lightweight,
parameter-efficient baseline model known for its stability on smaller EEG
datasets. \textbf{ShallowConvNet} and
\textbf{DeepConvNet}~\citep{Schirrmeister2017DeepConv} were selected as
spatial baselines for their proven ability to extract oscillatory energy patterns and build deep
hierarchical features across the temporal and
spatial domains. \textbf{CNN--LSTM} merges convolutional spatial
extraction with recurrent temporal memory, enabling modeling of
sequential dynamics at the cost of increased training complexity.
\textbf{TCN}~\citep{Ingolfsson2020TCNet} explicitly models long-range
temporal dependencies using dilated causal convolutions while maintaining
stable optimization. \textbf{GNN--CNN}~\citep{Song2020GNN} explores
inter-channel functional connectivity through graph convolutions using a
predefined spatial adjacency matrix based on the 128-channel EEG montage.

Table~\ref{tab:model_comparison} provides a detailed architectural
comparison of all six models.

\begin{table*}[t]
\centering
\caption{Architectural trade-offs and performance characteristics of evaluated deep learning models}\label{tab:model_comparison}
\begin{tabular}{@{}p{0.14\textwidth} p{0.18\textwidth} p{0.18\textwidth} p{0.1\textwidth} p{0.18\textwidth} p{0.14\textwidth}@{}}
\toprule
\textbf{Model} & \textbf{Spatial Modeling} & \textbf{Temporal Modeling} & \textbf{Complexity} & \textbf{Core Strength} & \textbf{Limitation} \\
\midrule
EEGNet         & Depthwise filters         & Short-range convolutions   & Low                 & Parameter-efficient    & Limited context \\
ShallowConvNet & Shallow convolutions      & Band-power filtering       & Low--Med            & Interpretable features & Insufficient depth \\
DeepConvNet    & Hierarchical features     & Multi-scale blocks         & High                & Strong accuracy        & Inter-subject variability \\
CNN--LSTM      & CNN-based encoder         & Recurrent modeling         & High                & Sequential dynamics    & Training instability \\
TCN            & Implicit integration      & Dilated causal convs       & Med--High           & Best trend alignment   & Slightly higher MAE \\
GNN--CNN       & Graph connectivity        & Temporal graph convs       & High                & Explicit relationships & Reduced generalization \\
\bottomrule
\end{tabular}
\end{table*}

\section{Experiments and Results}\label{sec6}

\subsection{Evaluation Protocol and Metrics}

All models were evaluated exclusively using the 15-train / 5-test
participant-wise split, preventing any subject-level information leakage.
Performance is assessed using the four metrics described in
Section 4: MAE, RMSE, Pearson~$r$, and tolerance accuracy.

\subsection{Regression Performance on MUSIN-G}

The results, detailed in Table~\ref{tab:results}, underscore the inherent
difficulty of cross-subject enjoyment prediction. Deep convolutional
models achieved the lowest absolute error (MAE $\approx 0.97$--0.99). On
the 1--5 Likert scale, this corresponds roughly to an adjacent ordinal
disagreement, which is expected in unpersonalized cross-subject tasks.
Temporal models showed a distinct complementary strength: the TCN achieved
the highest Pearson correlation ($r = 0.21$), demonstrating its superior
ability to capture relative enjoyment trends as a song progresses.

\begin{table}[ht]
\centering
\caption{Regression performance on the MUSIN-G dataset (15-train / 5-test participant-wise split)}\label{tab:results}
\small
\setlength{\tabcolsep}{2pt}
\begin{tabular}{@{}lcccc@{}}
\toprule
\textbf{Model} & \textbf{MAE}$\downarrow$ & \textbf{RMSE}$\downarrow$ & \textbf{\shortstack{Pearson\\$r\uparrow$}} & \textbf{\shortstack{Tol. Acc.\\(\%)$\uparrow$}} \\
\midrule
EEGNet         & 1.03 & 1.27 & $-0.01$       & 46.7 \\
ShallowConvNet & 1.31 & 1.56 & $-0.06$       & 35.0 \\
DeepConvNet    & 0.99 & 1.24 & 0.20          & 51.7 \\
CNN--LSTM      & 0.98 & \textbf{1.18} & 0.18 & 56.7 \\
TCN            & 1.05 & 1.28 & \textbf{0.21} & 55.0 \\
GNN--CNN       & \textbf{0.97} & 1.19 & 0.18 & \textbf{58.3} \\
\bottomrule
\end{tabular}
\end{table}

Table~\ref{tab:bestmodel} summarizes which model excels on each criterion.

\begin{table}[ht]
\centering
\caption{Best-performing model per evaluation criterion}\label{tab:bestmodel}
\small
\setlength{\tabcolsep}{3pt}
\begin{tabular}{@{}lll@{}}
\toprule
\textbf{Criterion} & \textbf{Best Model} & \textbf{Observation} \\
\midrule
Lowest MAE         & GNN--CNN            & Best absolute accuracy \\
Lowest RMSE        & CNN--LSTM           & Best against large errors \\
Highest Pearson $r$& TCN                 & Best trend alignment \\
Highest Tol.\ Acc. & GNN--CNN            & Best proximity \\
\bottomrule
\end{tabular}
\end{table}

\subsection{Trend Alignment and Explainability Analysis}

To verify the statistical validity of the TCN trend alignment, we
performed permutation testing under the strict cross-subject split, which
yielded a $p$-value of $0.056$ (indicating marginal significance). Given the strict cross-subject split protocol and the unpersonalized nature of the task, a $p$-value of $0.056$ provides strong evidence of trend preservation that warrants future validation on larger cohorts. Furthermore, pairwise Wilcoxon tests comparing the MAE of DeepConvNet
and TCN showed no statistically significant difference,
reinforcing that spatial and temporal architectures capture highly complementary
aspects of enjoyment.

\begin{figure}[ht]
\centering
\includegraphics[width=\columnwidth]{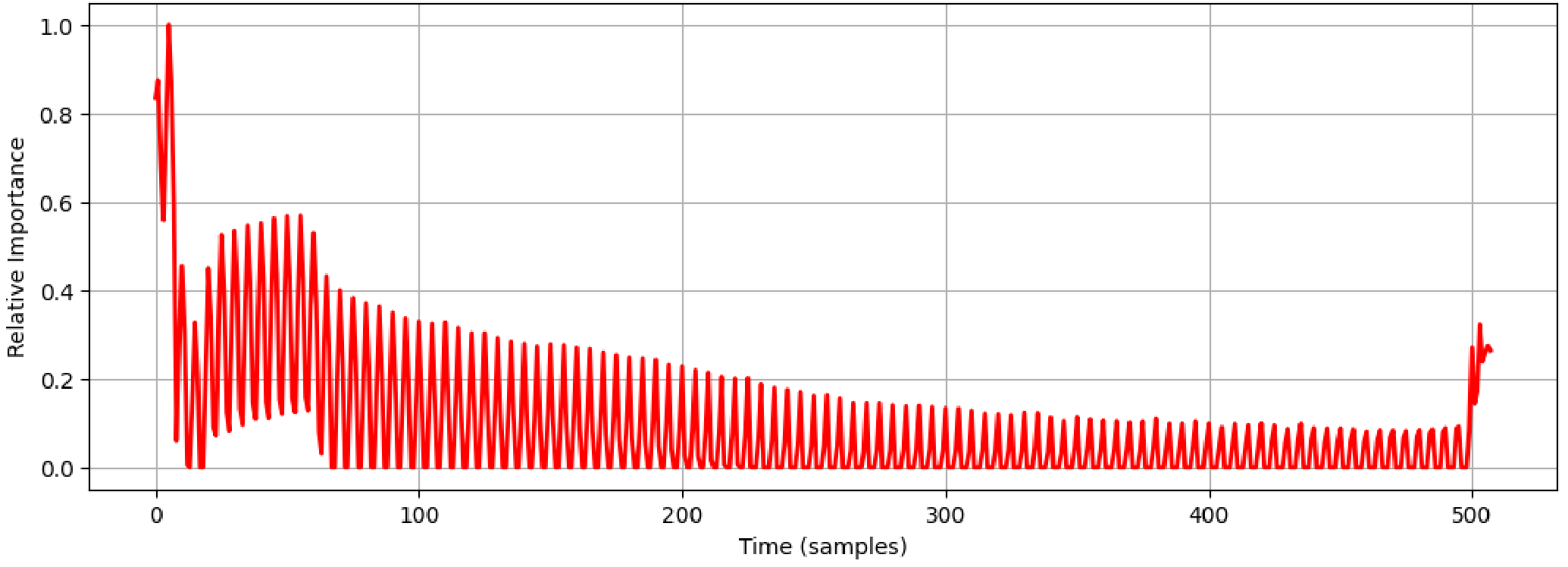}
\caption{Grad-CAM visualization highlighting temporally salient EEG
regions for enjoyment prediction. Relevance is highest at the start of the window and decays over time}
\label{fig:gradcam}
\end{figure}

\begin{figure}[ht]
\centering
\includegraphics[width=\columnwidth]{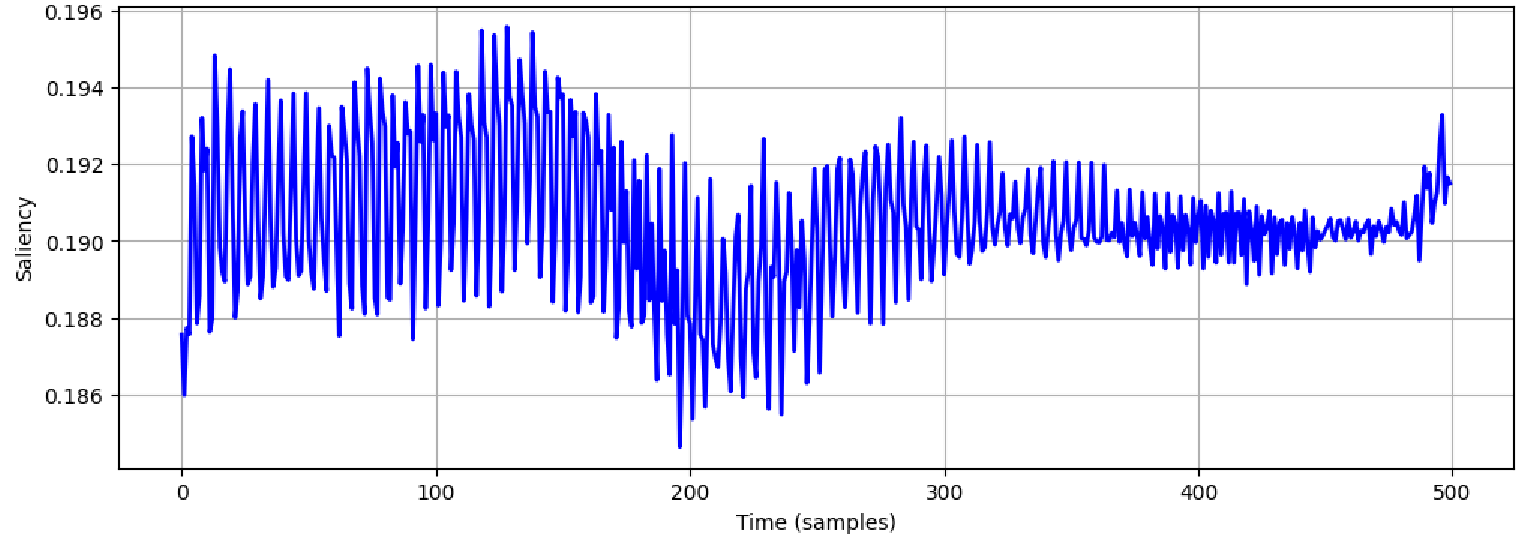}
\caption{Input gradient saliency map illustrating sensitivity of TCN
predictions to localized EEG signal variations across time.}
\label{fig:gradient_temporal}
\end{figure}

\begin{figure}[ht]
\centering
\includegraphics[width=\columnwidth]{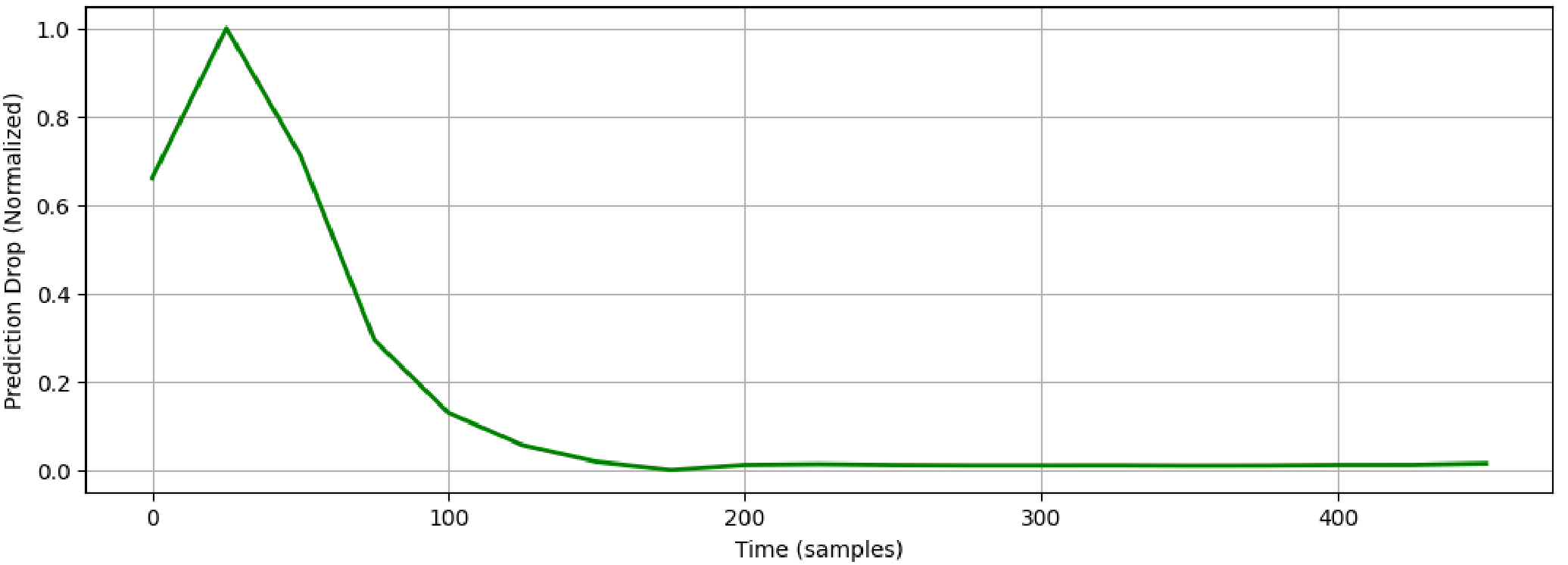}
\caption{Occlusion-based explainability analysis showing the contribution of
individual temporal EEG windows to enjoyment prediction}
\label{fig:occlusion_temporal}
\end{figure}

\begin{figure}[ht]
\centering
\includegraphics[width=\columnwidth]{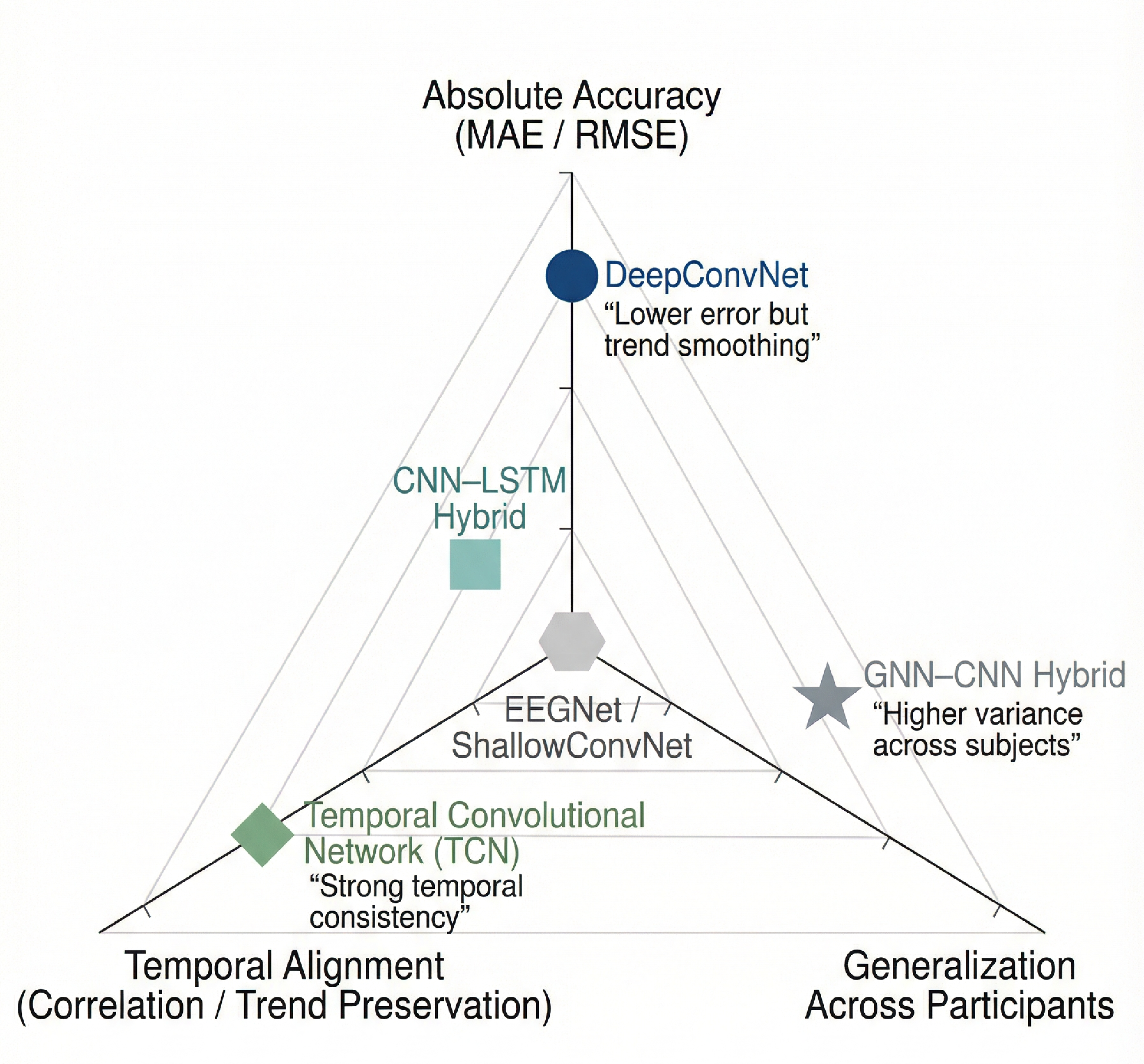}
\caption{Summary of XAI analysis. Temporal relevance is distributed
across sustained EEG segments rather than isolated events, supporting a
context-driven interpretation of enjoyment prediction}
\label{fig:xai}
\end{figure}

Three XAI methods were applied post-training to the
TCN~\citep{Sturm2016Interpretable}: gradient-based activation mapping
(Grad-CAM) \citep{Selvaraju2017GradCAM}, input gradient sensitivity, and occlusion-based explainability
analysis, with the resulting visualizations presented in Figure~\ref{fig:gradcam}, Figure~\ref{fig:gradient_temporal}, Figure~\ref{fig:occlusion_temporal}, and Figure~\ref{fig:xai}. These analyses confirm that the model's predictions do not rely
on isolated neural spikes but instead depend on distributed temporal EEG
dynamics, supporting the neuroscientific view that music enjoyment is a
sustained, evolving cognitive state.

\section{Discussion and Limitations}\label{sec7}

Our findings confirm that EEG signals contain measurable, contextually
rich information regarding subjective music enjoyment, aligning with
recent advances in affective
computing~\citep{Wang2023Survey,Ma2024Review}. The superior trend
alignment of temporal architectures underscores the need to treat music
enjoyment as an evolving sequence rather than a static state. The XAI
analyses further support this by showing that prediction quality depends
on accumulated temporal context rather than any single brief
neural event, consistent with neurophysiological accounts of music
perception~\citep{Madsen2019Synchrony}.

The negative Pearson correlation for EEGNet and ShallowConvNet suggests
that these compact models lack the representational capacity needed to
track relative enjoyment trends, even when their absolute errors are
moderate. The GNN--CNN model, despite achieving the best tolerance
accuracy, shows a limited positive correlation, suggesting that
graph-based representations may be too subject-specific to transfer
reliably across participants without personalization.

Several limitations must be acknowledged. The relatively small
participant pool in MUSIN-G constrains statistical power. Song-level
labels applied to two-second windows introduce unavoidable weak-label
noise, as human enjoyment naturally fluctuates within a piece. Relying
on EEG alone fails to account for contextual factors such as cultural
background, musical familiarity, and autobiographical memory. To mitigate the effects of weak-label noise, future iterations will explore multi-instance learning (MIL) paradigms or frame-level attention gating mechanisms to dynamically weight the relevance of individual temporal windows rather than utilizing uniform label assignment. Future work
should incorporate larger cohorts, multimodal integration, and
personalized fine-tuning to improve absolute predictive performance.

\section{Conclusion}\label{sec8}

This study presents a rigorous, regression-based and explainability-aware
framework for predicting music enjoyment from EEG data. By systematically
benchmarking six deep learning architectures on the MUSIN-G dataset under
a strict participant-wise protocol, we demonstrate that EEG signals
support meaningful cross-subject enjoyment prediction, while no single
architecture simultaneously minimizes absolute error and maximizes trend
alignment. Deep convolutional models minimize absolute error, while TCN
best captures relative enjoyment trends. Integrated XAI analysis
confirms that predictions rely on distributed temporal EEG dynamics,
providing a transparent foundation for future work in personalized,
neurologically driven affective computing.

\backmatter

\section*{Acknowledgments}

The authors sincerely thank the creators of the MUSIN-G dataset for making the dataset publicly available to the research community. The authors also acknowledge the support and academic environment provided by the Department of Computer Science and Engineering, Thapar Institute of Engineering and Technology, Patiala, India.

\section*{Declarations}

\textbf{Funding:}
No funding was received to assist with the preparation of this manuscript.

\noindent\textbf{Competing interests:}
The authors have no competing interests to declare that are relevant to the content of this article.

\noindent\textbf{Ethics approval:}
Not applicable. This study utilizes publicly available, anonymized datasets.

\noindent\textbf{Consent to participate:}
Not applicable.

\noindent\textbf{Consent for publication:}
Not applicable.

\noindent\textbf{Availability of data and materials:}
The MUSIN-G dataset analyzed during the current study is available in the OpenNeuro repository (https://doi.org/10.18112/openneuro.ds003774.v1.0.2).

\noindent\textbf{Code availability:}
Custom code developed for this study is available from the corresponding author upon reasonable request.

\noindent\textbf{Authors' contributions:}
Parshav Singla conceived the study, developed the methodology, implemented the deep learning models, conducted the experiments, analyzed the results, and prepared the original manuscript draft. Simranjit Kaur supervised the research, provided scientific guidance throughout the study, validated the methodology and findings, and critically reviewed and revised the manuscript. Maanya Walia contributed to literature survey, experimental validation, data analysis, and manuscript editing. Niharika contributed to result interpretation, performance evaluation, manuscript revision, proofreading, and overall technical discussion. All authors read and approved the final manuscript.

\end{document}